\documentclass[10pt,twocolumn,letterpaper]{article}
\usepackage{cvpr}
\pdfoutput=1
\usepackage{times}
\usepackage{epsfig}
\usepackage{graphicx}
\usepackage{amsmath}
\usepackage{amssymb}
\usepackage{subcaption}
\usepackage{float}
\usepackage{multirow}
\usepackage{enumitem}
\usepackage{autobreak}
\usepackage[breaklinks=true,bookmarks=false]{hyperref}
\cvprfinalcopy 
\def\cvprPaperID{****} 

\begin{document}
\allowdisplaybreaks
\title{Future Frame Prediction of a Video Sequence}
\author{Jasmeen Kaur\\
Visualization \& Perception Laboratory\\Dept. of CS\&E, IIT Madras\\
Chennai, $600036$, India\\
{\tt\small jasmeenes23@gmail.com}
\and
Sukhendu Das\\
Visualization \& Perception Laboratory\\Dept. of CS\&E, IIT Madras\\
Chennai, $600036$, India\\
{\tt\small sdas@cse.iitm.ac.in}
}
\maketitle
\begin{abstract}
\par Predicting future frames of a video sequence has been a problem of high interest in the field of Computer Vision as it caters to a multitude of applications. The ability to predict, anticipate and reason about future events is the essence of intelligence and one of the main goals of decision-making systems such as human-machine interaction, robot navigation and autonomous driving. However, the challenge lies in the ambiguous nature of the problem as there may be multiple future sequences possible for the same input video shot. A naively designed model averages multiple possible futures into a single blurry prediction.
\par Recently, two distinct approaches have attempted to address this problem as: (a) use of latent variable models that represent underlying stochasticity and (b) adversarially trained models that aim to produce sharper images. A latent variable model often struggles to produce realistic results, while an adversarially trained model underutilizes latent variables and thus fails to produce diverse predictions. These methods have revealed complementary strengths and weaknesses. Combining the two approaches produces predictions that appear more realistic and better cover the range of plausible futures. This forms the basis and objective of study in this project work. 
\par In this paper, we proposed a novel multi-scale architecture combining both approaches. We validate our proposed model through a series of experiments and empirical evaluations on Moving MNIST, UCF101, and Penn Action datasets. Our method outperforms the results obtained using the baseline methods. 
\end{abstract}

\section{Introduction}
\par Future event anticipation is one of the key components of any intelligent decision-making system. This task of predicting events from a given video clip, has recently attracted the attention of key research communities in the fields of Computer Vision and Machine/Deep Learning.
\par Although this task is relatively easy for humans, it presents a wide array of challenges from a machine’s perspective. Many distracting artifacts, such as occlusions, changing camera angles, lighting conditions, object deformations or clutter further complicate the task. Despite such challenges, a lot of methods have been able to generate outputs with a certain degree of success.
\par
\begin{figure}[htbp]
\begin{center}
\includegraphics[scale=0.16]{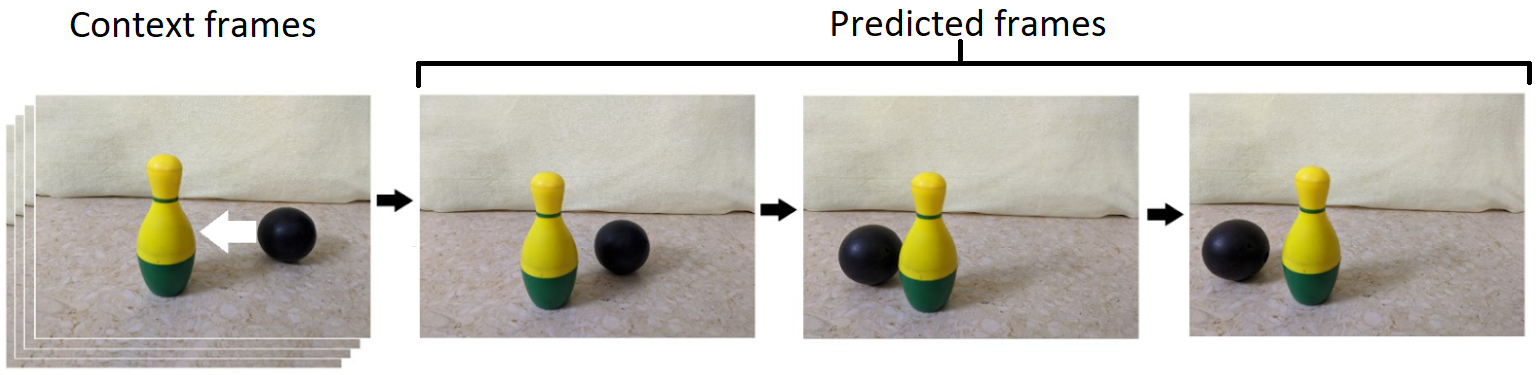}
\caption{A bowling ball moving towards a bowling pin. Our aim is to predict whether there will be a collision or not. Prediction of such collisions is very important in the context of autonomous cars.}
\label{example}
\end{center}
\end{figure}

 Viewing a sequence of images allows humans to make a remarkable number of judgements about the future. Lets consider an example. See the frames of a short video clip in Figure \ref{example}. Will the bowling ball hit the pin? This might be the first question that crosses our minds after observing Figure \ref{example}. Finding an answer to this question might be a difficult task but if we observe the sequence of images carefully, we will observe certain clues that can aid us in forecasting the future, e.g., from the viewer’s perspective the bowling ball is farther than the pin. So it would roll away, passing from behind the pin and consequently avoiding the collision. This is just one among the several scenarios where predicting future frames of the video can be useful. 
\par As is apparent from this example, vital cues for the task of tracking is optical flow field/vectors to forecast future impending action, expected in the scene. Most published work \cite{wei18, li18} have thus used such features or their variants such as – SIFT/HOG correspondences, spatio-temporal gradients, LBPTOPs \cite{hong16}, etc. Post tracking (dense or sparse), most of the methods apply optimization criteria or GAN-based deep learning methods to forecast the future. In addition, producing crispy clear future frames at a long distance (time) in the future has also been of interest to researchers.    
\par The rest of the paper is organized as follows. Section 2 describes the problem definition. Section 3 details the multiple approaches used for anticipating future frames of a video. Section 4 outlines the proposed technique and the objective function used. Finally, Section 5 introduces three real world datasets used for evaluation. We also show the qualitative and quantitative results of our experiments and Section 6 concludes the paper. 
\section{Problem Definition}
\par Video prediction is the task of predicting subsequent frames, given a sequence of video frames. Formally, it can be defined as follows. Let $\mathbf{X_t} \in R^{w \times h \times c}$ be the $t$-th frame in the input video sequence $\mathbf{X} = (X_{t-n}, \dots, X_{t-1}, X_{t})$ consisting of $n$ frames. Here, $w$, $h$ and $c$ denote the width, height and number of channels respectively. The target is to predict the subsequent $m$ frames $\mathbf{Y} = (\hat{Y}_{t+1}, \hat{Y}_{t+2}, \dots, \hat{Y}_{t+m})$ with crisp clarity and high precision, using the input $\mathbf{X}$.
\par The problem is not as elementary as it may appear at first glance. It has been observed that the frames in the future become more blurred with increasing time stamps in output $Y$, and reduction of frames in given input $X$. Also, prediction of the future is a multi-modal problem, i.e, the same past may have many different possible futures. Even if a model correctly learns hidden representations and understands motion details, it may predict a future different from the actual sequence. Varying patterns of  motion of objects in video clips over a large variety of scenarios may be difficult to learn by a machine. If video has clutter, large set of objects in motion, less clarity (blur, low illumination etc.)
the task becomes harder.
\par In this work, we assume that the input video clip does not have jitter or any sharp change of scene. There is not much of aliasing or blurring artifacts. The moving object(s) is/are not camouflaged by the other objects or background. The moving objects do not follow a random motion path. Training and testing datasets are from similar sensors and acquisition environments, with limited variations across domains.
\section{Related Works}
There is a wide spectrum of methodologies and approaches pertaining to the task of video prediction.\\
\textbf{Direct Pixel Synthesis:}
Early video prediction models did not take into account the explicit modeling of scene dynamics and directly attempted to predict future pixel intensities. Ranzato \etal \cite{ranzato14} introduced a method wherein video frames were discretized into patch clusters. It employs the K-means algorithm to form clusters.\\
\textbf{Using Explicit Transformations:}
Some authors have pointed out that trying to hallucinate the whole frame from scratch results in blurry predictions. They argue that in most of the cases, a large part of a frame is unchanged or is a result of small patch displacements. In other words, they propose that a model should be able to account only for the differences of the future frame with respect to the past sequence. Then those differences can be applied to the last input in order to predict the next frame. Using this, an architecture should be better able to predict as it has to account for less variability. Klein \etal \cite{klein15} and Brabandere \etal \cite{brab16} use a model that predicts transformations in the form of filters which are then applied to the last input frame. Finn \etal \cite{finn16} predict multiple transformations for separate patches and then a dynamic mask is applied for giving different weight to different patches. \\
\textbf{In High-level Feature Space:}
In recent years, much attention has shifted to high-level representations which have emerged as a promising avenue for complete scene understanding. Video prediction methods deal with the curse of dimensionality by reducing the prediction space to high-level representations,such as human pose, semantic segmentation, and instance segmentation. High-level prediction spaces constitute good intermediate representations and are more tractable. Models that bypass the predictions in the pixel space are capable of reporting more accurate and long term predictions. Bhattacharjee \etal \cite{bhattacharjee19} proposed a Graph Convolution Network (GCN) based architecture for semantic segmentation of futuristic frames. Bhattacharjee \etal \cite{bhattacharjee18} presented feature reconstruction based approach using GANs.\\
\textbf{Explicit Separation of Content and Motion:}
Seeking inspiration from two-stream architectures used in action recognition \cite{simonyan14}, authors tried to implement a similar paradigm in video prediction context. They tried to process content and motion on separate pathways by factorizing the video. High dimensional videos are decomposed to perform prediction on lower-dimensional temporal dynamics independent of the spatial layout. Better results have been achieved by factorizing the task of prediction into more tractable task even though this makes end-to-end training difficult. The first end-to-end model capable of isolating the visual appearance from scene dynamics was the Motion-content Network (MCnet) \cite{vill17}. \\
\textbf{Conditioned on Extra Variables:}
Some architectures use some extra features along with the input video, i.e. the future frames are conditioned on actions or states. Conditioning the prediction on extra variables such as vehicle odometry, robot state, etc. narrows the prediction space. The dynamics of the scene are directly affected by these variables and thus they are able to provide beneficial data that facilitates the task of prediction. For instance, acceleration and wheel-steering influence the motion of an autonomous vehicle, captured by a camera placed on the dashboard of the autonomous vehicle. In the absence of such information, we still rely on the model’s competence to correlate the acceleration and wheel-steering with the perceived motion. Therefore, using these variables explicitly guides the prediction and is highly advantageous. The above method was exploited by Oh \etal \cite{oh15}, who pioneered the model for long term video predictions in Atari games based on next action choosen by the player. \\
\textbf{Incorporating Uncertainty:}
For a single input video sequence, multiple outcomes are equally probable i.e. the underlying distribution has many modes. Classification and regression approaches intend to discretize a high dimensional continuous space and regress to the mean respectively, thus addressing multimodal distribution is not straightforward for them. For dealing with the intrinsic unpredictability of videos, authors tried to incorporate uncertainty in prediction models. This was done either by relying on generative models like GANs and VAEs or by introducing latent variables into existing deterministic models. \eg, Bhattacharjee \etal \cite{bhattacharjee17} proposed the use of a multi-stage (2-stage) generative adversarial network. Goroshin \etal \cite{goroshin15} introduced latent variables into a convolutional AE and proposed a probabilistic netork capable of linearly extrapolating predicted frame in a feature space by learning linearized feature representations. 
\section{Proposed Architecture}
The proposed architecture uses SAVP architecture proposed by Lee \etal \cite{savp} as the baseline architecture. The architecture is modified so that it is able to predict future for more time stamps and also alleviate mode collapse problem. 
\subsection{SAVP Architecture}
\par The SAVP architecture as proposed in "Stochastic Adversarial Video Prediction" \cite{savp} combines GAN and VAE. GANs are trained with randomly drawn input, whereas VAEs only observe ground truth images and are never trained on random input which may lead to a train-test mismatch. Thus, SAVP model takes advantage of their complementary strengths.
\subsection{Modifications Proposed}
\begin{itemize}
    \item \textbf{Replace Convolutional LSTM with E3D-LSTM \cite{eidetic}:} Convolutional LSTM predicts future frame based on sequentially updated memory states. When the memory cell is refreshed, older memories are discarded. In contrast, the proposed E3D-LSTM model \cite{eidetic} maintains a list of historical memory records and revokes them when necessary, thereby facilitating long-range video reasoning. The Eidetic 3D LSTM (E3D-LSTM) has a new state, namely RECALL state which can recall long term past unlike LSTM. RECALL mechanism helps to recall temporally distant memory.
    \item \textbf{Replace GAN with MG-GAN \cite{mggan}:} Generative adversarial networks are susceptible to mode collapse. For alleviating mode collapse, Manifold Guided Generative Adversarial Network (MGGAN) \cite{mggan} is used. MGGAN leverages a guidance network on the existing GAN architecture. The guidance network consists of an encoder and a discriminator. In order to solve the mode collapse, we design the encoder such that the output distribution of encoder is the best approximate of $p_{data}$. So, it captures all modes of data.  To this end, we adopt an encoder of a pre-trained autoencoder as a manifold mapping function. Because autoencoder aims to reconstruct all samples in dataset, the encoder should not ignore minor modes . Consequently, the generator avoids mode missing during training because it receives feedback for minor modes of data distribution from the guidance network.
\end{itemize}
\subsection{Modified SAVP Architecture}
\par
\begin{figure}[htbp]
\centering
\includegraphics[width=\linewidth, height=0.5\linewidth]{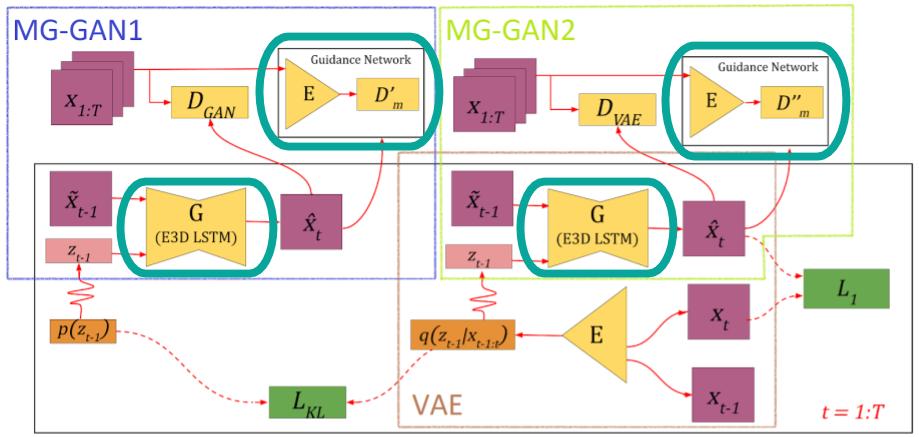}
\caption{Modified SAVP Architecture}
\label{savp_modified}
\end{figure}

 The modified SAVP architecture is shown in Figure \ref{savp_modified}. The modifications made are encircled. The architecture has two MGGANs \cite{mggan} and a VAE. The decoder of the VAE also as a generator. MGGAN integrates a guidance network to the existing GAN architecture. The guidance network encourages the generator to learn all modes of $p_{data}$. The guidance network's goal is to teach generator that $p_{model}$ and $p_{data}$ match in the learned manifold space. The guidance network comprises of an encoder for manifold mapping and a discriminator for measuring dissimilarity between the distributions of $p_{data}$ and $p_{model}$ in the manifold space. To cope with mode collapse, guidance network uses an encoder layer of a pre-trained autoencoder. This autoencoder is optimized for reconstructing all samples of real images, and is kept fixed after pre-training so that it does not get updated during the training of GAN. 
\subsection{Multi-Scale Architecture}
\par
\begin{figure}[htbp]
\centering
\includegraphics[width=1.1\linewidth, height=0.5\linewidth]{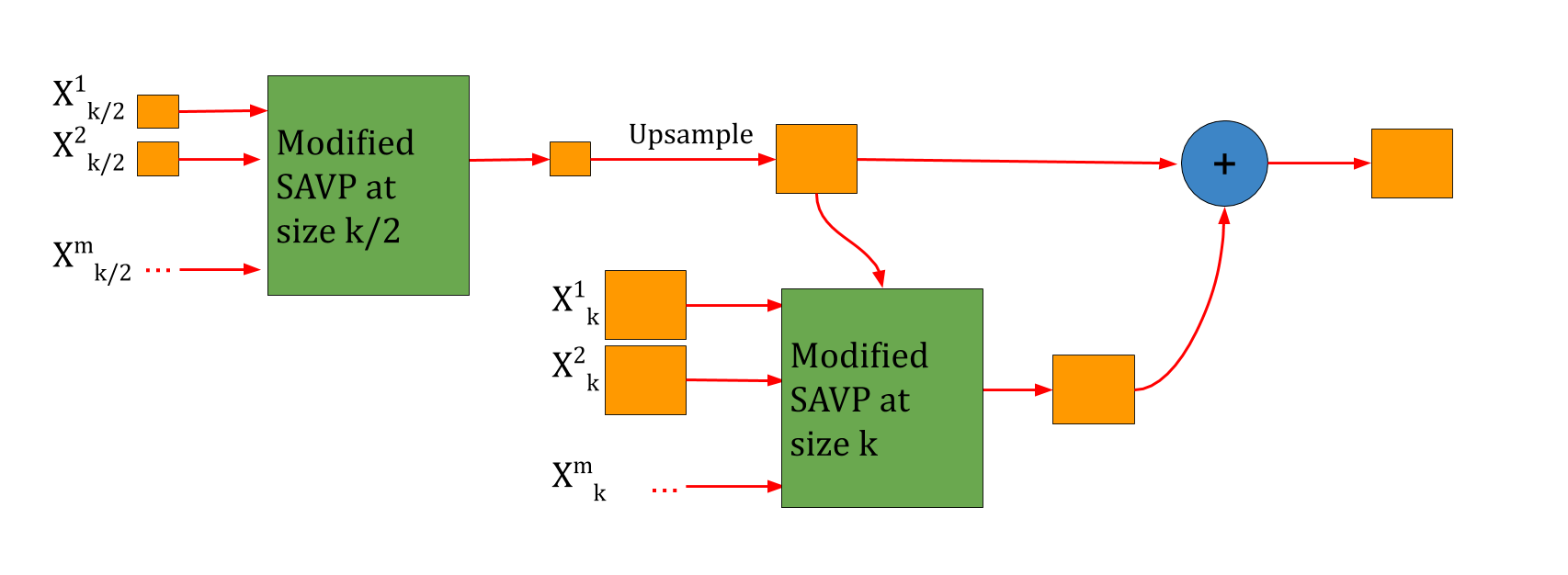}
\caption{Multiscale Architecture, inspired from \cite{mathieu16}.}
\label{multiscale}
\end{figure}

 Since convolutions only account for short-range relationships, we use pooling layers for garnering information from a wider range. However, this technique leads to the generation of low resolution images. To overcome this, \cite{mathieu16} suggested the use of multi-scale architecture to combine multiple scales linearly as shown in Figure \ref{multiscale}.
\subsection{Objective Function}
\par The objective function used is similar to the objective function used by Lee \etal \cite{savp}]for SAVP model. The GAN losses are replaced by the MGGAN losses \cite{mggan}.\\ \\
\textbf{L1 Loss} The L1 loss between the predicted image and the ground truth is minimised.
\begin{equation}
    \begin{aligned}
        L_1(G,E) = \mathbb E_{x_{0:T},z_t \sim E\left(x_{t:t+1}\right)|_{t=0}^{T-1}}\left[\sum_{t=1}^{T}\left|\left|x_t - G\left(x_0,z_{0,t-1}\right)\right|\right|_1\right]  
    \end{aligned}
\end{equation}
\textbf{KL Divergence Loss} The KL Divergence loss between the prior distribution $p\left(z_{t-1}\right)$ and the posterior distribution $\mathbb E\left(x_{t-1:t}\right)$ is minimised.
\begin{equation}
    \begin{aligned}
        L_{KL}\left(E\right) = \mathbb E_{x_{0:T}}\left[\sum_{t=1}^{T}D_{KL}\left(\mathbb E\left(x_{t-1:t}\right)||p\left(z_{t-1}\right)\right)\right]
    \end{aligned}
\end{equation}
\textbf{MGGAN Loss} The following functions show the objective function of MGGANs.
\begin{equation}
\begin{aligned}
L_{MGGAN1} &= \mathbb E_{x_{1:T}}[\log ( D(x_{0:T-1})) + \log ( D_{m}'( \\ & E(x_{0:T-1})))] + \mathbb E_{x_{1:T},z_t \sim p(z_t)|_{t=0}^{T-1}}[\log \\ & (1 - D(G(x_0,z_{0:T-1}))) + \log(1 - D_{m}' \\ & (E(G(x_0,z_{0:T-1}))))] 
\end{aligned}
\end{equation}
\begin{equation}
\begin{aligned}
L_{MGGAN2} &= \mathbb E_{x_{1:T}}[\log ( D^{VAE}(x_{0:T-1})) + \log ( D_{m}'' \\ &( E(x_{0:T-1})))] + \mathbb E_{x_{1:T},z_t \sim p(z_t)|_{t=0}^{T-1}}[\log(1 \\ & - D^{VAE}(G(x_0,z_{0:T-1}))) +  \log(1 - D_{m}'' \\ & (E(G(x_0,z_{0:T-1}))))]
\end{aligned}
\end{equation}
\textbf{Combined Loss} Finally, we combine different losses with different weights as: 
\begin{equation}
\begin{aligned}
L_{combined} = \lambda_{1}L_{1} + \lambda_{2}L_{KL} + \lambda_{3}L_{MGGAN1} +\lambda_{4}L_{MGGAN2},
\end{aligned}
\end{equation}
$\lambda_{i}'s; i=\{1, \dots, 4\}$ are set empirically as $\lambda_{1}=0.25, \lambda_{2}=0.2, \lambda_{3}=0.3$ and $\lambda_{4}=0.3.$
\section{Experiments}
Performance analysis with experiments of our proposed prediction model for video frame(s) have been done on video clips from UCF-101 \cite{ucf101}, Moving MNIST \cite{mnist} and Penn Action \cite{penn} datasets. The input-output configuration used for training the system is as follows:
\begin{itemize}
\item Input: 10 frames
\item Output: 10 and 20 frames.
\end{itemize}
\subsection{Datasets}
The datasets used for performance evaluation are: \\
\begin{itemize}
    \item \textbf{UCF101 Dataset \cite{ucf101}:} This dataset contains $13320$ annotated action videos. The videos have been accumulated from YouTube and has $101$ different categories of action. The videos have a resolution of $320 \times 240$ pixels and a frame rate of $25$ fps. $10000$ videos are used for training and $3320$ videos for testing.
    \item \textbf{Moving MNIST Dataset \cite{mnist}:} This dataset has randomly sampled two digits from the original MNIST dataset, floating and bouncing at boundaries at constant velocity and angle inside a $64 \times 64$ patch. New sequences can be generated as and when required making the dataset an almost infinite source of video sequences. $8000$ videos are used for training and $2000$ videos for testing.
    \item \textbf{Penn Action Dataset \cite{penn}:} This dataset comprises $2326$ video sequences which have been collected from the University of Pennsylvania. It contains videos from $15$ different categories of actions. The videos have been annotated for human joint as well as camera viewpoint (i.e. the camera position with reference to a human). $2000$ videos are used for training and $326$ videos for testing.
\end{itemize} 
\subsection{Performance Evaluation}
\textbf{Quantitative Results}
\par The quality of the predicted frames is assessed using three metrics: \\
(a) Mean-squared Error (MSE)\\
(b) Peak Signal to Noise Ratio (PSNR) \\
(c) Structural Similarity Index Measure (SSIM). \\
"PSNR measures the quality of the reconstruction process through the calculation of Mean-squared error between the original and the reconstructed signal in logarithmic decibel scale" \cite{bovik09}. "SSIM is also an image similarity measure where, one of the images being compared is assumed to be of perfect quality" \cite{ssim}. SSIM ranges between $-1$ and $1$ where a larger score indicates greater similarity between the two images. \\
$x \longrightarrow y$ represents the task of taking $x$ context frames as input and predicting next $y$ frames. \\
The results are shown in Tables \ref{ucf}, \ref{mnist} and \ref{penn}. We have represented the best results in bold. \\
\begin{table}[H]
    \centering
    \begin{tabular}{llll}
            \hline
            \multirow{2}{*}{\textsc{Model}} & \multicolumn{3}{c}{\textbf{10 $\longrightarrow{}$ 10}}  \\
                                &\multicolumn{1}{l}{MSE}
                                 &\multicolumn{1}{l}{PSNR}   &\multicolumn{1}{l}{SSIM}  \\
            \hline 
            E3D-LSTM \cite{eidetic} &47.6 &25.34 &0.82 \\
            SAVP \cite{savp} &52.3 &19.61 &0.78  \\
            \textsc{Modified} SAVP &45.1 &27.42 &0.84  \\
            \textsc{Ours} &\textbf{43.8} &\textbf{29.58} &\textbf{0.86} \\
            \hline \\
            \hline
            \multirow{2}{*}{\textsc{Model}} & \multicolumn{3}{c}{\textbf{10 $\longrightarrow{}$ 20}} \\
                                & \multicolumn{1}{l}{MSE}
                                 &\multicolumn{1}{l}{PSNR}  &\multicolumn{1}{l}{SSIM} \\
            \hline 
            E3D-LSTM \cite{eidetic} &53.6 &21.46 &0.74 \\
            SAVP \cite{savp} &56.5 &17.34 &0.71 \\
            \textsc{Modified} SAVP &51.4 &24.57 &0.76 \\
            \textsc{Ours} &\textbf{49.2} &\textbf{28.24} &\textbf{0.79} \\
            \hline
        \end{tabular}
        \caption{Quantitative Results on UCF101 dataset on 10 $\longrightarrow{}$ 10 and 10 $\longrightarrow{}$ 20.}
    \label{ucf}
\end{table}
\begin{table}[H]
    \centering
    \begin{tabular}{llll}
            \hline
            \multirow{2}{*}{\textsc{Model}} & \multicolumn{3}{c}{\textbf{10 $\longrightarrow{}$ 10}} \\
                                &\multicolumn{1}{l}{MSE}
                                 &\multicolumn{1}{l}{PSNR}   &\multicolumn{1}{l}{SSIM}  \\
            \hline 
            E3D-LSTM \cite{eidetic} &41.3 &29.41 &0.89  \\
            SAVP \cite{savp} &47.2 &21.26 &0.83  \\
            \textsc{Modified} SAVP &38.9 &31.55 &0.91  \\
            \textsc{Ours} &\textbf{33.5} &\textbf{35.67} &\textbf{0.95}  \\
            \hline \\
            \hline
            \multirow{2}{*}{\textsc{Model}} & \multicolumn{3}{c}{\textbf{10 $\longrightarrow{}$ 20}} \\
                                 & \multicolumn{1}{l}{MSE}
                                 &\multicolumn{1}{l}{PSNR}  &\multicolumn{1}{l}{SSIM} \\
            \hline 
            E3D-LSTM \cite{eidetic} &46.9 &23.62 &0.81 \\
            SAVP \cite{savp} &50.3 &20.52 &0.75 \\
            \textsc{Modified} SAVP &39.7 &26.75 &0.83 \\
            \textsc{Ours} &\textbf{41.2} &\textbf{30.24} &\textbf{0.87} \\
            \hline
        \end{tabular}
        \caption{Quantitative Results on Moving MNIST dataset on 10 $\longrightarrow{}$ 10 and 10 $\longrightarrow{}$ 20.}
    \label{mnist}
\end{table}
\begin{table}[H]
    \centering
    \begin{tabular}{llll}
            \hline
            \multirow{2}{*}{\textsc{Model}} & \multicolumn{3}{c}{\textbf{10 $\longrightarrow$ 10}} \\
                                &\multicolumn{1}{l}{MSE}
                                 &\multicolumn{1}{l}{PSNR}   &\multicolumn{1}{l}{SSIM}  \\
            \hline 
            E3D-LSTM \cite{eidetic} &44.6 &26.25 &0.84 \\
            SAVP \cite{savp} &49.7 &19.73 &0.80 \\
            \textsc{Modified} SAVP &41.2 &28.46 &0.86 \\
            \textsc{Ours} &\textbf{38.6} &\textbf{29.19} &\textbf{0.89} \\
            \hline \\
            \hline
            \multirow{2}{*}{\textsc{Model}} & \multicolumn{3}{c}{\textbf{10 $\longrightarrow$ 20}} \\
                                 & \multicolumn{1}{l}{MSE}
                                 &\multicolumn{1}{l}{PSNR}  &\multicolumn{1}{l}{SSIM} \\
            \hline 
            E3D-LSTM \cite{eidetic} &48.6 &21.92 &0.78 \\
            SAVP \cite{savp} &53.5 &18.29 &0.72 \\
            \textsc{Modified} SAVP &45.7 &25.78 &0.81 \\
            \textsc{Ours} &\textbf{43.8} &\textbf{26.34} &\textbf{0.85} \\
            \hline
        \end{tabular}
        \caption{Quantitative Results on Penn Action dataset on 10 $\longrightarrow{}$ 10 and 10 $\longrightarrow{}$ 20.}
    \label{penn}
\end{table}
From Table \ref{ucf}, Table \ref{mnist} and Table \ref{penn}, we can infer that the proposed architecture outperforms the baseline methods.\\ \\
\textbf{Qualitative Results}
\par
\begin{figure*}[htbp]
\vspace{-1.8cm}
\begin{subfigure}[b]{\textwidth}
\includegraphics[width=\textwidth,height=0.45\textwidth]{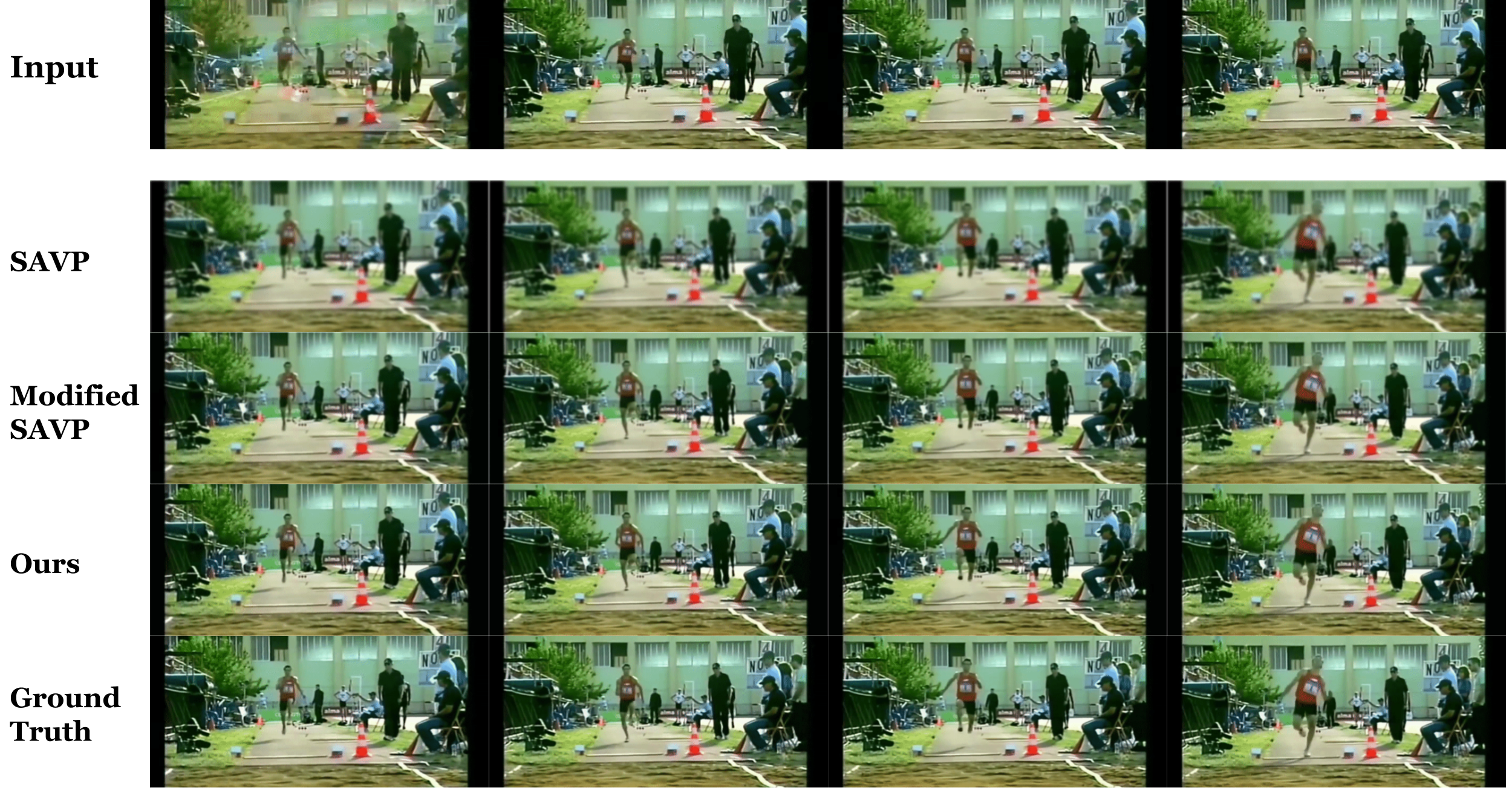}
\end{subfigure}
\begin{subfigure}[b]{\textwidth}
\includegraphics[width=\textwidth,height=0.45\textwidth]{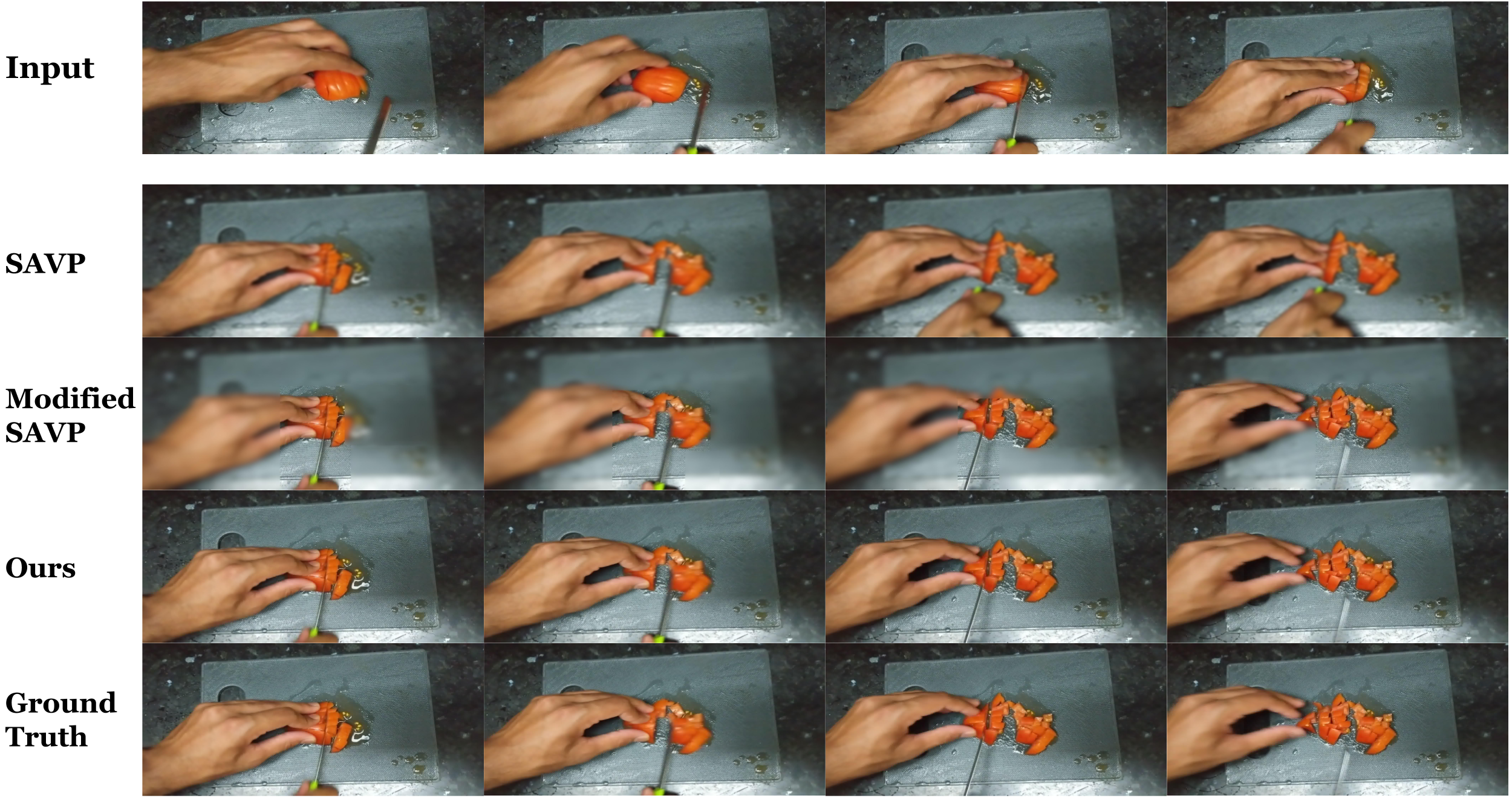}
\end{subfigure}
\begin{subfigure}[b]{\textwidth}
\includegraphics[width=\textwidth,height=0.45\textwidth]{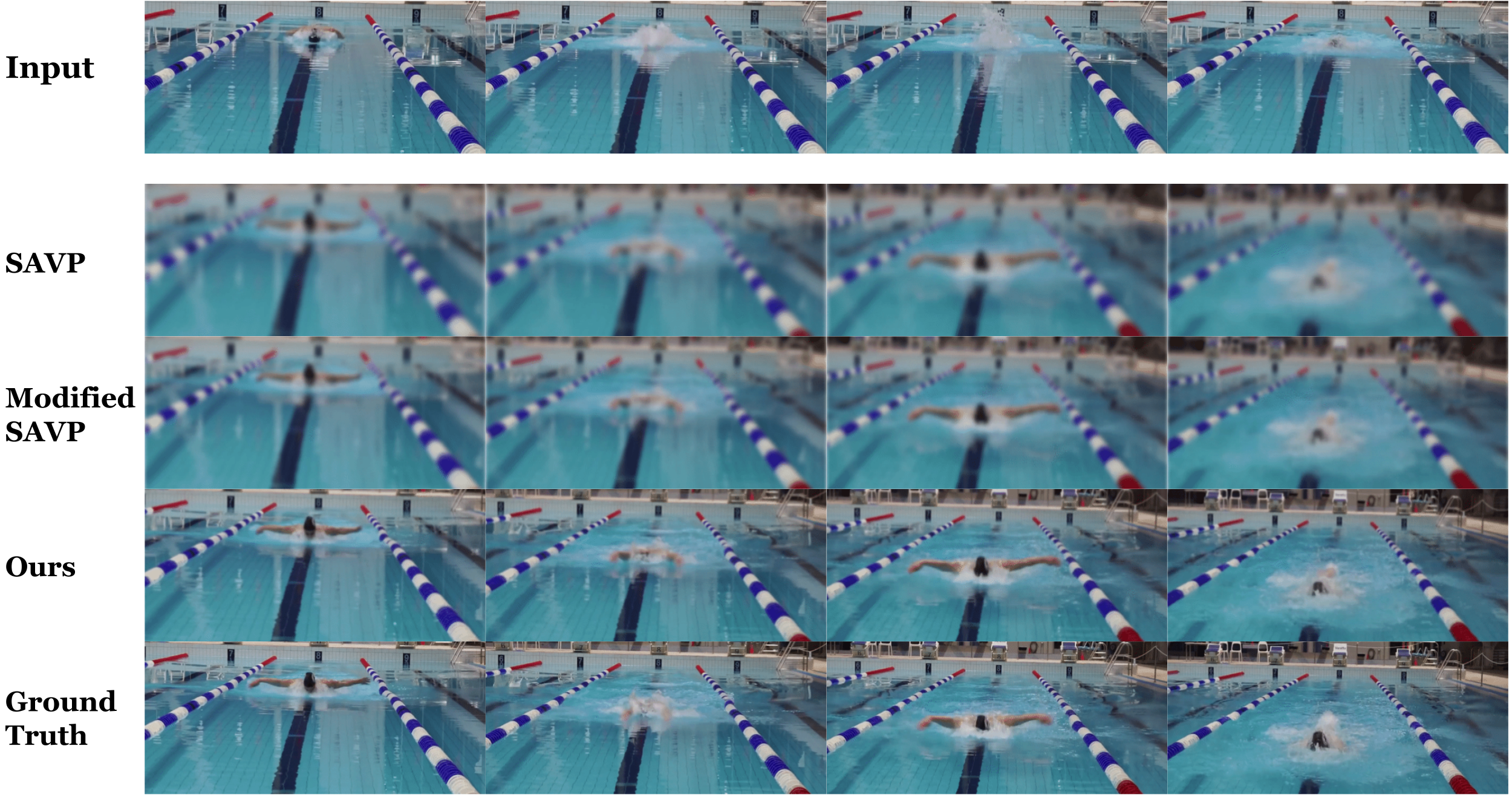}
\end{subfigure}
\caption{Qualitative results with four rows representing SAVP, Modified SAVP, Proposed Architecture (Ours) results and Ground Truth.}
\label{results}
\end{figure*}

 The qualitative results for frame prediction by using the proposed framework are shown in Figure \ref{results}. The four rows in Figure \ref{results} represent 
\begin{enumerate}
\item SAVP results
\item Modified SAVP results
\item Proposed Architecture (Ours) results
\item Ground-truth
\end{enumerate}
The output of the proposed multi-scale architecture is closest to the ground truth, as exhibited by the last rows of Figure \ref{results}. 
\par The video results are given on \url{https://drive.google.com/drive/folders/15iyYJEYjnMewTzXCHADEOHNV9pp2TTHs?usp=sharing}.
\section{Conclusion}
\par In this paper, we have attempted to combine the strengths of the two approaches used in "Stochastic Adversarial Video Prediction" \cite{savp} and "Eidetic 3D LSTM: A Model for Video Prediction and Beyond" \cite{eidetic} and proposed a novel architecture for the task of video prediction. We have compared four models: SAVP, E3D-LSTM, ModifiedSAVP, and our Proposed Architecture quantitatively and qualitatively on three datasets: UCF101, Moving MNIST, and Penn Action Dataset. We observed that our proposed Architecture outperformed the base models by a significant margin. Future scope of work involves testing of models on synthetically generated datasets and using alternative evaluation metrics.

{\small
\bibliographystyle{ieee}
\bibliography{egbib}
}

\end{document}